\documentclass[10pt,twocolumn,letterpaper]{article}

\usepackage{cvpr}
\usepackage{times}
\usepackage{epsfig}
\usepackage{graphicx}
\usepackage{amsmath}
\usepackage{amssymb}

\usepackage{url}            
\usepackage{booktabs}       
\usepackage{amsfonts}       
\usepackage{nicefrac}       
\usepackage{microtype}      
\usepackage{algorithm}
\usepackage{algpseudocode}
\usepackage{graphicx}
\usepackage{amsmath}
\usepackage{multirow}
\usepackage{color}
\usepackage{subfigure}

\newcommand{\tabincell}[2]{\begin{tabular}{@{}#1@{}}#2\end{tabular}}

\usepackage[breaklinks=true,bookmarks=false]{hyperref}

\cvprfinalcopy 

\def\cvprPaperID{377} 

\begin{document}

\title{GhostNet: More Features from Cheap Operations}

\author{Kai Han$^1$\quad Yunhe Wang$^1$\quad Qi Tian$^{1}$\thanks{Corresponding author}\quad Jianyuan Guo$^2$\quad Chunjing Xu$^1$\quad Chang Xu$^3$\\
$^1$Noah's Ark Lab, Huawei Technologies.\qquad  $^2$Peking University.\\
$^3$School of Computer Science, Faculty of Engineering, University of Sydney.\\
{\tt\small \{kai.han,yunhe.wang,tian.qi1,xuchunjing\}@huawei.com jyguo@pku.edu.cn c.xu@sydney.edu.au}
}

\maketitle

\begin{abstract}
Deploying convolutional neural networks (CNNs) on embedded devices is difficult due to the limited memory and computation resources. The redundancy in feature maps is an important characteristic of those successful CNNs, but has rarely been investigated in neural architecture design. This paper proposes a novel Ghost module to generate more feature maps from cheap operations. Based on a set of intrinsic feature maps, we apply a series of linear transformations with cheap cost to generate many ghost feature maps that could fully reveal information underlying intrinsic features. The proposed Ghost module can be taken as a plug-and-play component to upgrade existing convolutional neural networks. Ghost bottlenecks are designed to stack Ghost modules, and then the lightweight GhostNet can be easily established. Experiments conducted on benchmarks demonstrate that the proposed Ghost module is an impressive alternative of convolution layers in baseline models, and our GhostNet can achieve higher recognition performance (\eg $75.7\%$ top-1 accuracy) than MobileNetV3 with similar computational cost on the ImageNet ILSVRC-2012 classification dataset. Code is available at \url{https://github.com/huawei-noah/ghostnet}.
\end{abstract}

\section{Introduction}
\begin{figure}[htp]
	\centering
	\includegraphics[width=0.9\linewidth]{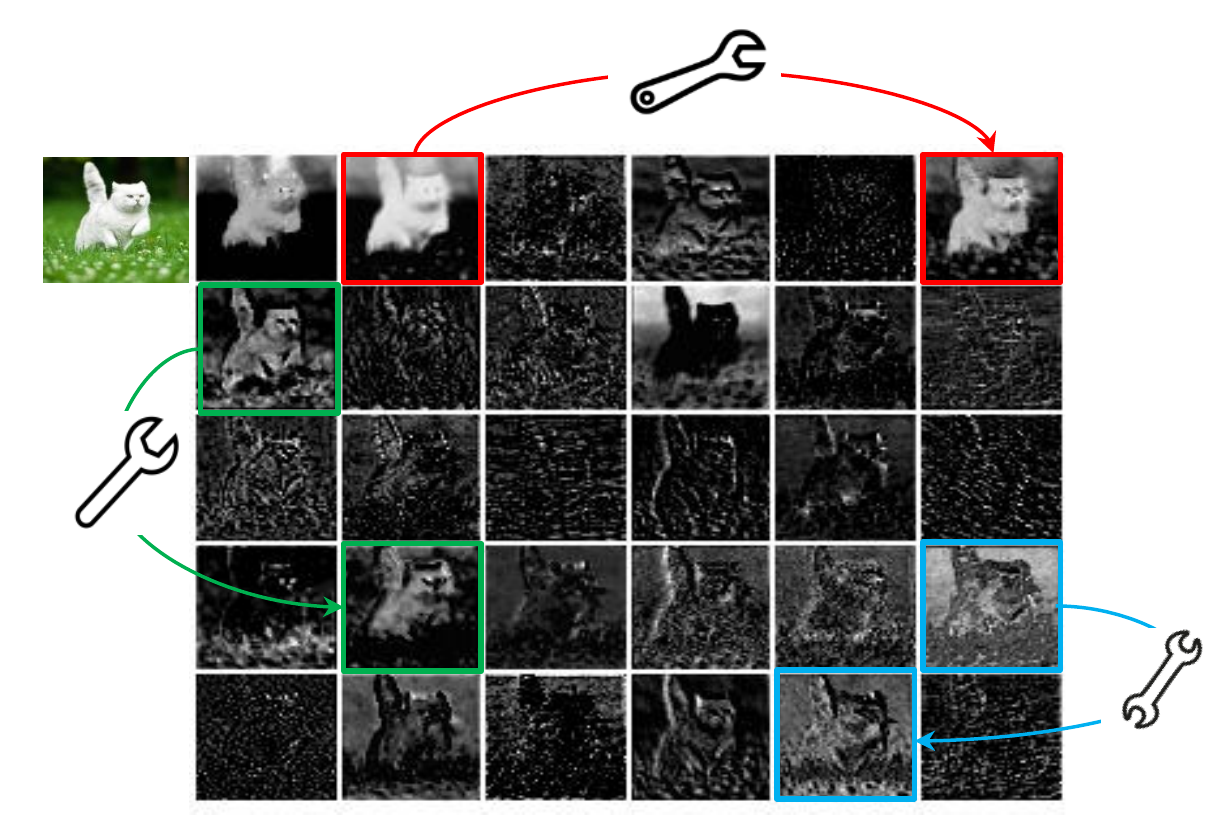}
	\caption{Visualization of some feature maps generated by the first residual group in ResNet-50, where three similar feature map pair examples are annotated with boxes of the same color. One feature map in the pair can be approximately obtained by transforming the other one through cheap operations (denoted by spanners).}
	\label{Fig:maps}
	\vspace{-1em}
\end{figure}

Deep convolutional neural networks have shown excellent performance on various computer vision tasks, such as image recognition~\cite{alexnet,a3m}, object detection~\cite{fasterrcnn,retinanet}, and semantic segmentation~\cite{deeplab}. Traditional CNNs usually need a large number of parameters and floating point operations (FLOPs) to achieve a satisfactory accuracy, \eg ResNet-50~\cite{resnet} has about $25.6$M parameters and requires $4.1$B FLOPs to process an image of size $224\times224$. Thus, the recent trend of deep neural network design is to explore portable and efficient network architectures with acceptable performance for mobile devices (\eg smart phones and self-driving cars).

Over the years, a series of methods have been proposed to investigate compact deep neural networks such as network pruning~\cite{deepcompression,thinet}, low-bit quantization~\cite{xnor,jacob2018quantization}, knowledge distillation~\cite{Distill,you2017learning}, \etc. Han~\etal~\cite{deepcompression} proposed to prune the unimportant weights in neural networks. \cite{l1-pruning} utilized $\ell_1$-norm regularization to prune filters for efficient CNNs. ~\cite{xnor} quantized the weights and the activations to 1-bit data for achieving large compression and speed-up ratios. \cite{Distill} introduced knowledge distillation for transferring knowledge from a larger model to a smaller model. However, performance of these methods are often upper bounded by pre-trained deep neural networks that have been taken as their baselines.

Besides them, efficient neural architecture design has a very high potential for establishing highly efficient deep networks with fewer parameters and calculations, and recently has achieved considerable success. This kind of methods can also provide new search unit for automatic search methods~\cite{rlnas,yang2019cars,chen2019fasterseg}. For instance, MobileNet~\cite{mobilenet,mobilev2,mobilenetv3} utilized the depthwise and pointwise convolutions to construct a unit for approximating the original convolutional layer with larger filters and achieved comparable performance. ShuffleNet~\cite{shufflenet,shufflev2} further explored a channel shuffle operation to enhance the performance of lightweight models.

Abundant and even redundant information in the feature maps of well-trained deep neural networks often guarantees a comprehensive understanding of the input data. For example, Figure~\ref{Fig:maps} presents some feature maps of an input image generated by ResNet-50, and there exist many similar pairs of feature maps, like a \emph{ghost} of each another. Redundancy in feature maps could be an important characteristic for a successful deep neural network. Instead of avoiding the redundant feature maps, we tend to embrace them, but in a cost-efficient way. 

In this paper, we introduce a novel Ghost module to generate more features by using fewer parameters. Specifically, an ordinary convolutional layer in deep neural networks will be split into two parts. The first part involves ordinary convolutions but their total number will be rigorously controlled. Given the intrinsic feature maps from the first part, a series of simple linear operations are then applied for generating more feature maps. Without changing the size of output feature map, the overall required number of parameters and computational complexities in this Ghost module have been decreased, compared with those in vanilla convolutional neural networks. Based on Ghost module, we establish an efficient neural architecture, namely, GhostNet. We first replace original convolutional layers in benchmark neural architectures to demonstrate the effectiveness of Ghost modules, and then verify the superiority of our GhostNets on several benchmark visual datasets. Experimental results show that, the proposed Ghost module is able to decrease computational costs of generic convolutional layer while preserving similar recognition performance, and GhostNets can surpass state-of-the-art efficient deep models such as MobileNetV3~\cite{mobilenetv3}, on various tasks with fast inference on mobile devices.

The rest of the paper is organized as follows: section~\ref{related-work} briefly concludes the related work in the area, followed by the proposed Ghost module and GhostNet in section~\ref{Approach}, the experiments and analysis in section~\ref{Experiments}, and finally, conclusion in section~\ref{Conclusion}.


\section{Related Work}\label{related-work}
Here we revisit the existing methods for lightening neural networks in two parts: model compression and compact model design.

\subsection{Model Compression}
For a given neural network, model compression aims to reduce the computation, energy and storage cost~\cite{deepcompression,wang2019e2,gui2019model,xu2019positive}. Pruning connections~\cite{han2015learning,deepcompression,cnnpack} cuts out the unimportant connections between neurons. Channel pruning~\cite{wen2016learning,he2017channel,l1-pruning,thinet,nisp,huang2018data,liu2019learning} further targets on removing useless channels for easier acceleration in practice. Model quantization~\cite{xnor,bnn,jacob2018quantization} represents weights or activations in neural networks with discrete values for compression and calculation acceleration. Specifically, binarization methods~\cite{bnn,xnor,bireal,shen2019searching} with only 1-bit values can extremely accelerate the model by efficient binary operations. Tensor decomposition~\cite{jaderberg2014speeding,denton2014exploiting} reduces the parameters or computation by exploiting the redundancy and low-rank property in weights. Knowledge distillation~\cite{Distill,han2018co,chen2019data} utilizes larger models to teach smaller ones, which improves the performance of smaller models. The performances of these methods usually depend on the given pre-trained models. The improvement on basic operations and architectures will make them go further.

\subsection{Compact Model Design}
With the need for deploying neural networks on embedded devices, a series of compact models are proposed in recent years~\cite{xception,mobilenet,mobilev2,mobilenetv3,shufflenet,shufflev2,shift,yang2019legonet}. Xception~\cite{xception} utilizes depthwise convolution operation for more efficient use of model parameters. MobileNets~\cite{mobilenet} are a series of light weight deep neural networks based on depthwise separable convolutions. MobileNetV2~\cite{mobilev2} proposes inverted residual block and MobileNetV3~\cite{mobilenetv3} further utilizes AutoML technology~\cite{rlnas,yang2019cars,gong2019autogan} achieving better performance with fewer FLOPs. ShuffleNet~\cite{shufflenet} introduces channel shuffle operation to improve the information flow exchange between channel groups. ShuffleNetV2~\cite{shufflev2} further considers the actual speed on target hardware for compact model design. Although these models obtain great performance with very few FLOPs, the correlation and redundancy between feature maps has never been well exploited.

\section{Approach}\label{Approach}
In this section, we will first introduce the Ghost module to utilize a few small filters to generate more feature maps from the original convolutional layer, and then develop a new GhostNet with an extremely efficient architecture and high performance.

\subsection{Ghost Module for More Features}
Deep convolutional neural networks~\cite{alexnet,vgg,resnet} often consist of a large number of convolutions that results in massive computational costs. Although recent works such as MobileNet~\cite{mobilenet,mobilev2} and ShuffleNet~\cite{shufflev2} have introduced depthwise convolution or shuffle operation to build efficient CNNs using smaller convolution filters (floating-number operations), the remaining $1\times1$ convolution layers would still occupy considerable memory and FLOPs.

Given the widely existing redundancy in intermediate feature maps calculated by mainstream CNNs as shown in Figure~\ref{Fig:maps}, we propose to reduce the required resources, \ie convolution filters used for generating them. In practice, given the input data $X\in\mathbb{R}^{c\times h\times w}$, where $c$ is the number of input channels and $h$ and $w$ are the height and width of the input data, respectively,  the operation of an arbitrary convolutional layer for producing $n$ feature maps can be formulated as
\begin{equation}\label{eq:conv}
Y = X*f+b,
\end{equation}
where $*$ is the convolution operation, $b$ is the bias term, $Y\in\mathbb{R}^{h'\times w'\times n}$ is the output feature map with $n$ channels, and $f\in\mathbb{R}^{c\times k\times k \times n}$ is the convolution filters in this layer. In addition, $h'$ and $w'$ are the height and width of the output data, and $k\times k$ is the kernel size of convolution filters $f$, respectively. During this convolution procedure, the required number of FLOPs can be calculated as $n\cdot h'\cdot w'\cdot c\cdot k\cdot k$, which is often as large as hundreds of thousands since the number of filters $n$ and the channel number $c$ are generally very large (\eg 256 or 512).


\begin{figure}[ht]
	\vspace{-0.5em}
	\centering 
	\subfigure[The convolutional layer.] {
		\includegraphics[width=0.5\linewidth]{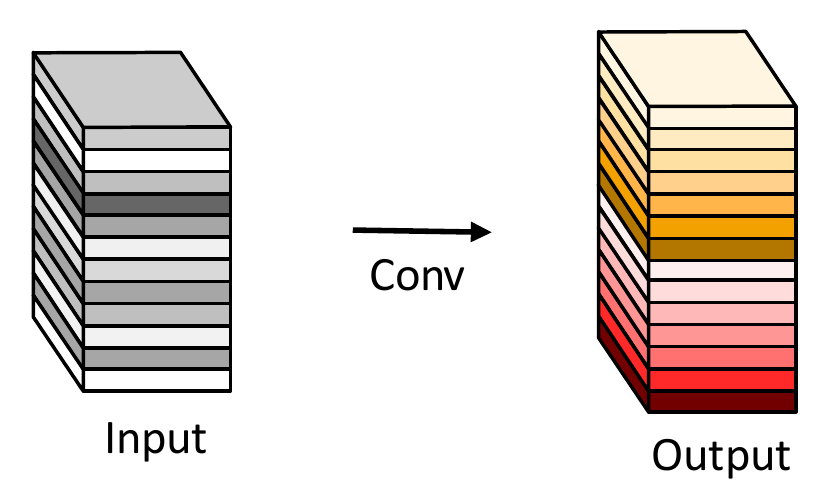}
	}
	\subfigure[The Ghost module.] {
		\includegraphics[width=0.9\linewidth]{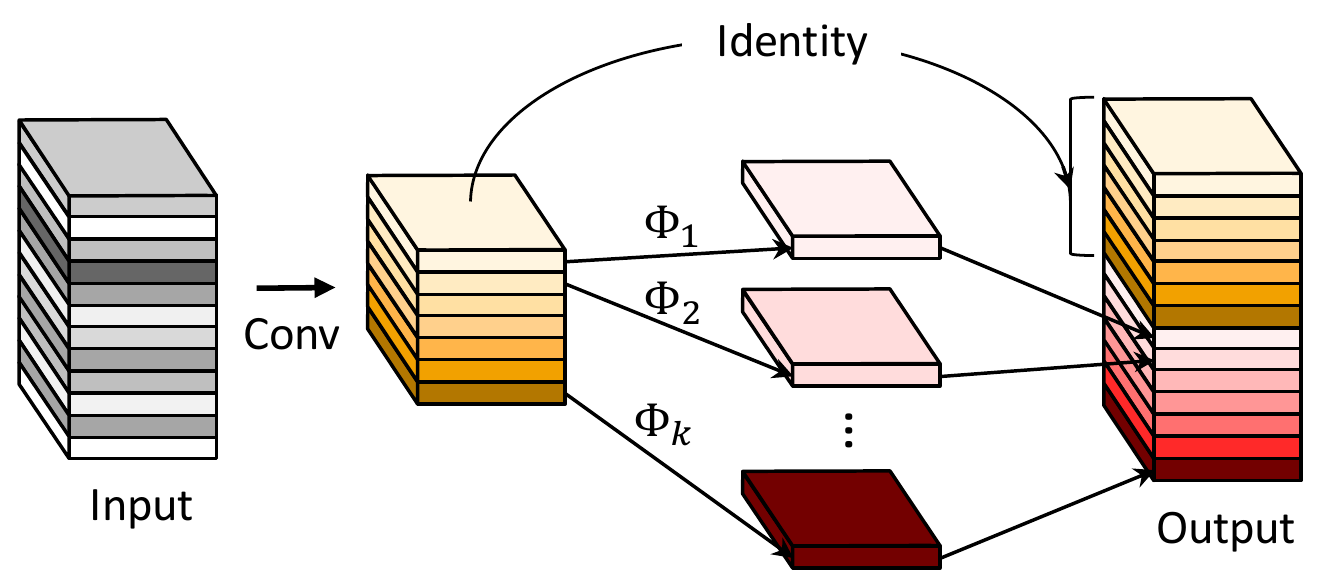} 
	}
	\caption{An illustration of the convolutional layer and the proposed Ghost module for outputting the same number of feature maps. $\Phi$ represents the cheap operation.} 
	\label{Fig:Ghost}
	\vspace{-0.5em}
\end{figure}

According to Eq.~\ref{eq:conv}, the number of parameters (in $f$ and $b$) to be optimized is explicitly determined by the dimensions of input and output feature maps. As observed in Figure~\ref{Fig:maps}, the output feature maps of convolutional layers often contain much redundancy, and some of them could be similar with each other. We point out that it is unnecessary to generate these redundant feature maps one by one with large number of FLOPs and parameters. Suppose that the output feature maps are ``ghosts'' of a handful of intrinsic feature maps with some cheap transformations. These intrinsic feature maps are often of smaller size and produced by ordinary convolution filters. Specifically, $m$ intrinsic feature maps $Y'\in\mathbb{R}^{h'\times w'\times m}$ are generated using a primary convolution:
\begin{equation}
Y' = X*f',
\end{equation}
where $f'\in\mathbb{R}^{c\times k\times k \times m}$ is the utilized filters, $m\leq n$ and the bias term is omitted for simplicity. The hyper-parameters such as filter size, stride, padding, are the same as those in the ordinary convolution (Eq.~\ref{eq:conv}) to keep the spatial size (\ie $h'$ and $w'$) of the output feature maps consistent. To further obtain the desired $n$ feature maps, we propose to apply a series of cheap linear operations on each intrinsic feature in $Y'$ to generate $s$ ghost features according to the following function:
\begin{equation}
y_{ij} = \Phi_{i,j}(y'_i),\quad \forall\; i = 1,...,m,\;\; j = 1,...,s,
\label{eq:ghost}
\end{equation}
where $y'_i$ is the $i$-th intrinsic feature map in $Y'$, $\Phi_{i,j}$ in the above function is the $j$-th (except the last one) linear operation for generating the $j$-th ghost feature map $y_{ij}$, that is to say, $y'_i$ can have one or more ghost feature maps $\{y_{ij}\}_{j=1}^{s}$. The last $\Phi_{i,s}$ is the identity mapping for preserving the intrinsic feature maps as shown in Figure~\ref{Fig:Ghost}(b). By utilizing Eq.~\ref{eq:ghost}, we can obtain $n=m\cdot s$ feature maps $Y=[y_{11},y_{12},\cdots,y_{ms}]$ as the output data of a Ghost module as shown in Figure~\ref{Fig:Ghost}(b). Note that the linear operations $\Phi$ operate on each channel whose computational cost is much less than the ordinary convolution. In practice, there could be several different linear operations in a Ghost module, \eg $3\times 3$ and $5\times5$ linear kernels, which will be analyzed in the experiment part.

\noindent\textbf{Difference from Existing  Methods.} The proposed Ghost module has major differences from existing efficient convolution schemes. i) Compared with the units in~\cite{mobilenet,shufflenet} which utilize $1\times 1$ pointwise convolution widely, the primary convolution in Ghost module can have customized kernel size. ii) Existing methods~\cite{mobilenet,mobilev2,shufflenet,shufflev2} adopt pointwise convolutions to process features across channels and then take depthwise convolution to process spatial information. In contrast, Ghost module adopts ordinary convolution to first generate a few intrinsic feature maps, and then utilizes cheap linear operations to augment the features and increase the channels. iii)  The operation to process each feature map is limited to depthwise convolution or shift operation in previous efficient architectures~\cite{mobilenet,shufflenet,shift,active-shift}, while linear operations in Ghost module can have large diversity. iv) In addition, the identity mapping is paralleled with linear transformations in Ghost module to preserve the intrinsic feature maps.

\noindent\textbf{Analysis on Complexities.} Since we can utilize the proposed Ghost module in Eq.~\ref{eq:ghost} to generate the same number of feature maps as that of an ordinary convolutional layer, we can easily integrate the Ghost module into existing well designed neural architectures to reduce the computation costs. Here we further analyze the profit on memory usage and theoretical speed-up by employing the Ghost module. For example, there are $1$ identity mapping and $m \cdot (s-1) = \frac{n}{s}\cdot (s-1)$ linear operations, and the averaged kernel size of each linear operation is equal to $d\times d$. Ideally, the $n\cdot(s-1)$ linear operations can have different shapes and parameters, but the online inference will be obstructed especially considering the utility of CPU or GPU cards. Thus, we suggest to take linear operations of the same size (\eg $3\times3$ or $5\times5$) in one Ghost module for efficient implementation. The theoretical speed-up ratio of upgrading ordinary convolution with the Ghost module is
\begin{equation}
\begin{aligned}
r_s &= \frac{n\cdot h'\cdot w'\cdot c\cdot k\cdot k}{\frac{n}{s}\cdot h'\cdot w'\cdot c\cdot k\cdot k + (s-1)\cdot \frac{n}{s}\cdot h'\cdot w'\cdot d\cdot d}\\
&= \frac{c\cdot k\cdot k}{\frac{1}{s}\cdot c\cdot k\cdot k+\frac{s-1}{s}\cdot d\cdot d} \approx \frac{s\cdot c}{s+c-1}\approx s,
\end{aligned}
\label{eq:rs}
\end{equation}
where $d\times d$ has the similar magnitude as that of $k\times k$, and $s\ll c$. Similarly, the compression ratio can be calculated as
\begin{equation}
\begin{aligned}
r_c &= \frac{n\cdot c\cdot k\cdot k}{\frac{n}{s}\cdot c\cdot k\cdot k + (s-1)\cdot\frac{n}{s}\cdot d\cdot d} \approx \frac{s\cdot c}{s+c-1} \approx s,
\end{aligned}
\label{eq:rc}
\end{equation}
which is equal to that of the speed-up ratio by utilizing the proposed Ghost module.

\begin{figure}[htb]
	\vspace{-1em}
	\centering
	\includegraphics[width=0.78\linewidth]{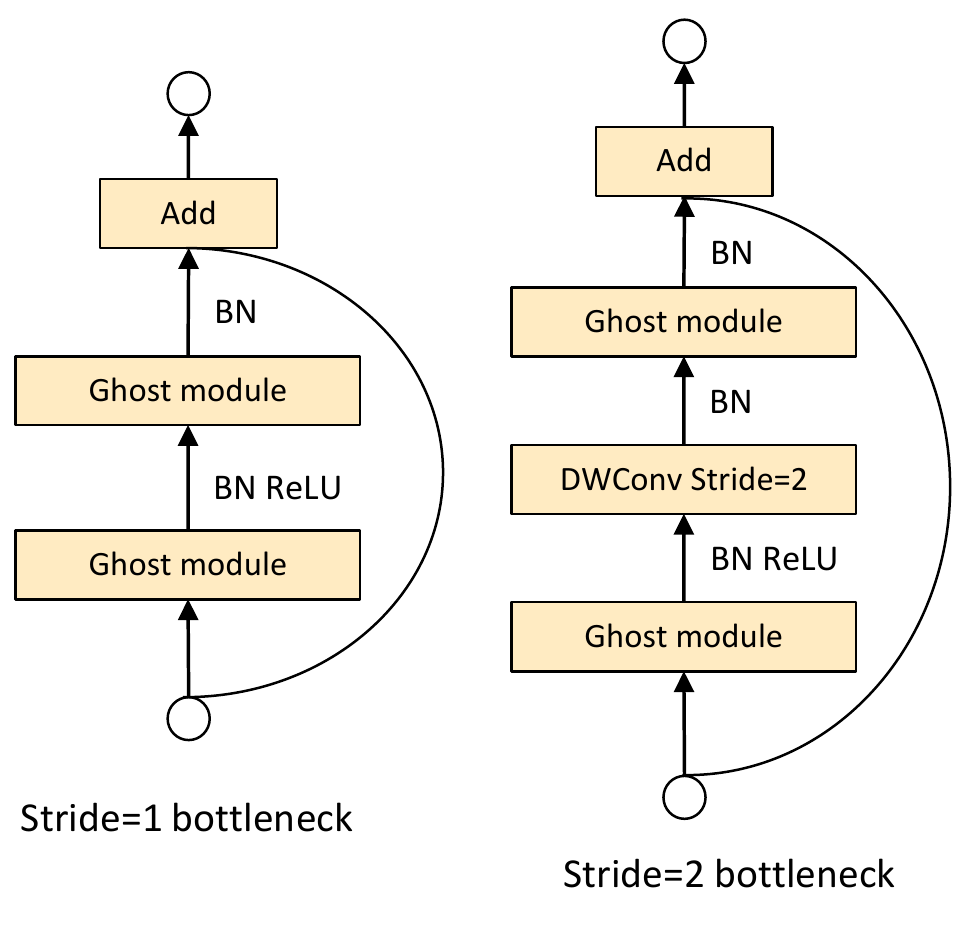}
	\vspace{-0.0em}
	\caption{Ghost bottleneck. Left: Ghost bottleneck with stride=1; right: Ghost bottleneck with stride=2.}
	\label{fig:hw}
	\vspace{-1em}
\end{figure}

\begin{table}[htb]
	\vspace{-0.1em}
	\centering
	\renewcommand\arraystretch{1.05}
	\caption{Overall architecture of GhostNet. G-bneck denotes Ghost bottleneck. \#exp means expansion size. \#out means the number of output channels. SE denotes whether using SE module.}
	\label{tab:GhostNet}
	\small
	\begin{tabular}{c||c|c|c|c|c}
		\hline
		Input & Operator & \#exp & \#out & SE & Stride \\ \hline\hline
		$224^2\times 3$   & Conv2d 3$\times$3 &  -  & 16     &  -  & 2 \\\hline
		$112^2\times 16$   & G-bneck &  16  & 16     &  -  & 1 \\
		$112^2\times 16$   & G-bneck &  48  & 24     &  -  & 2 \\\hline
		$56^2\times 24$   & G-bneck &  72  & 24     &  -  & 1 \\
		$56^2\times 24$   & G-bneck &  72  & 40      &  1  & 2 \\\hline
		$28^2\times 40$   & G-bneck &  120  & 40     &  1  & 1 \\
		$28^2\times 40$   & G-bneck &  240  & 80     &  -  & 2 \\\hline
		$14^2\times 80$   & G-bneck &  200  & 80     &  -  & 1 \\
		$14^2\times 80$   & G-bneck &  184  & 80    &  -  & 1 \\
		$14^2\times 80$   & G-bneck &  184  & 80    &  -  & 1 \\
		$14^2\times 80$   & G-bneck &  480  & 112     &  1  & 1 \\
		$14^2\times 112$   & G-bneck &  672  & 112     &  1  & 1 \\
		$14^2\times 112$   & G-bneck &  672  & 160     &  1  & 2 \\\hline
		$7^2\times 160$   & G-bneck &  960  & 160      &  -  & 1 \\
		$7^2\times 160$   & G-bneck &  960  & 160      &  1  & 1 \\
		$7^2\times 160$   & G-bneck &  960  & 160      &  -  & 1 \\
		$7^2\times 160$   & G-bneck &  960  & 160      &  1  & 1 \\
		$7^2\times 160$   & Conv2d 1$\times$1 &  - & 960     &  -  & 1 \\\hline
		$7^2\times 960$   & AvgPool 7$\times$7 &  -  & -      &  -  & - \\\hline
		$1^2\times 960$   & Conv2d 1$\times$1 &  -  & 1280     &  -  & 1 \\\hline
		$1^2\times 1280$   & FC &  -  & 1000     &  -  & - \\\hline
		\hline
	\end{tabular}
	\vspace{-1.0em}
\end{table}

\subsection{Building Efficient CNNs}
\paragraph{Ghost Bottlenecks.}
Taking the advantages of Ghost module, we introduce the Ghost bottleneck (G-bneck) specially designed for small CNNs. As shown in Figure~\ref{fig:hw}, the Ghost bottleneck appears to be similar to the basic residual block in ResNet~\cite{resnet} in which several convolutional layers and shortcuts are integrated. The proposed ghost bottleneck mainly consists of two stacked Ghost modules. The first Ghost module acts as an expansion layer increasing the number of channels. We refer the ratio between the number of the output channels and that of the input as \emph{expansion ratio}. The second Ghost module reduces the number of channels to match the shortcut path. Then the shortcut is connected between the inputs and the outputs of these two Ghost modules. The batch normalization (BN)~\cite{bn} and ReLU nonlinearity are applied after each layer, except that ReLU is not used after the second Ghost module as suggested by MobileNetV2~\cite{mobilev2}. The Ghost bottleneck described above is for stride=1. As for the case where stride=2, the shortcut path is implemented by a downsampling layer and a depthwise convolution with stride=2 is inserted between the two Ghost modules. In practice, the primary convolution in Ghost module here is pointwise convolution for its efficiency.

\paragraph{GhostNet.}

Building on the ghost bottleneck, we propose GhostNet as presented in Table~\ref{tab:GhostNet}. We basically follow the architecture of MobileNetV3~\cite{mobilenetv3} for its superiority and replace the bottleneck block in MobileNetV3 with our Ghost bottleneck. GhostNet mainly consists of a stack of Ghost bottlenecks with the Ghost modules as the building block. The first layer is a standard convolutional layer with 16 filters, then a series of Ghost bottlenecks with gradually increased channels are followed. These Ghost bottlenecks are grouped into different stages according to the sizes of their input feature maps. All the Ghost bottlenecks are applied with stride=1 except that the last one in each stage is with stride=2. At last a global average pooling and a convolutional layer are utilized to transform the feature maps to a 1280-dimensional feature vector for final classification. The squeeze and excite (SE) module~\cite{senet} is also applied to the residual layer in some ghost bottlenecks as in Table~\ref{tab:GhostNet}. In contrast to MobileNetV3, we do not use hard-swish nonlinearity function due to its large latency. The presented architecture provides a basic design for reference, although further hyper-parameters tuning or automatic architecture searching based ghost module will further boost the performance.

\paragraph{Width Multiplier.}
Although the given architecture in Table~\ref{tab:GhostNet} can already provide low latency and guaranteed accuracy, in some scenarios we may require smaller and faster models or higher accuracy on specific tasks. To customize the network for the desired needs, we can simply multiply a factor $\alpha$ on the number of channels uniformly at each layer. This factor $\alpha$ is called \emph{width multiplier} as it can change the width of the entire network. We denote GhostNet with width multiplier $\alpha$ as GhostNet-$\alpha\times$. Width multiplier can control the model size and the computational cost quadratically by roughly $\alpha^2$. Usually smaller $\alpha$ leads to lower latency and lower performance, and vice versa.

\section{Experiments}\label{Experiments}
In this section, we first replace the original convolutional layers by the proposed Ghost module to verify its effectiveness. Then, the GhostNet architecture built using the new module will be further tested on the image classification and object detection benchmarks.

\paragraph{Datasets and Settings.} To verify the effectiveness of the proposed Ghost module and GhostNet architecture, we conduct experiments on several benchmark visual datasets, including CIFAR-10~\cite{cifar}, ImageNet ILSVRC 2012 dataset~\cite{imagenet}, and MS COCO object detection benchmark~\cite{coco}. 

CIFAR-10 dataset is utilized for analyzing the properties of the proposed method, which consists of 60,000 $32\times32$ color images in 10 classes, with 50,000 training images and 10,000 test images. A common data augmentation scheme including random crop and mirroring~\cite{resnet,he2017channel} is adopted. ImageNet is a large-scale image dataset which contains over 1.2$M$ training images and 50$K$ validation images belonging to 1,000 classes. The common data preprocessing strategy including random crop and flip~\cite{resnet} is applied during training. We also conduct object detection experiments on MS COCO dataset~\cite{coco}. Following common practice~\cite{fpn,retinanet}, we train models on COCO \emph{trainval35k} split (union of 80$K$ training images and a random 35$K$ subset of images from validation set) and evaluate on the \emph{minival} split with 5$K$ images.


\subsection{Efficiency of Ghost Module}

\subsubsection{Toy Experiments.} We have presented a diagram in Figure~\ref{Fig:maps} to point out that there are some similar feature map pairs, which can be efficiently generated using some efficient linear operations. Here we first conduct a toy experiment to observe the reconstruction error between raw feature maps and the generated ghost feature maps. Taking three pairs in Figure~\ref{Fig:maps} (\ie red, greed, and blue) as examples, features are extracted using the first residual block of ResNet-50~\cite{resnet}. Taking the feature on the left as input and the other one as output, we utilize a small depthwise convolution filter to learn the mapping, \ie the linear operation $\Phi$ between them. The size of the convolution filter $d$ is ranged from $1$ to $7$, MSE (mean squared error) values of each pair with different $d$ are shown in Table~\ref{tab:toy}.

\begin{table}[htb]
	\centering
	\small
	\renewcommand\arraystretch{1.05}
	\caption{MSE error \emph{v.s.} different kernel sizes.}
	\vspace{0.1em}
	\label{tab:toy}
	\begin{tabular}{c||c|c|c|c}
		\hline
		MSE ($10^{-3}$) & $d$=1 & $d$=3 & $d$=5 & $d$=7 \\ \hline\hline
		red pair &  $4.0$    &  $3.3$    & $3.3$ &  $3.2$    \\ \hline
		green pair &  $25.0$     &  $24.3$    & $24.1$ &  $23.9$    \\ \hline
		blue pair &  $12.1$     &  $11.2$    & $11.1$ &  $11.0$    \\ \hline
	\end{tabular}
\end{table}

It can be found in Table~\ref{tab:toy} that all the MSE values are extremely small, which demonstrates that there are strong correlations between feature maps in deep neural networks and these redundant feature maps could be generated from several intrinsic feature maps. Besides convolutions used in the above experiments, we can also explore some other low-cost linear operations to construct the Ghost module such as affine transformation and wavelet transformation. However, convolution is an efficient operation already well support by current hardware and it can cover a number of widely used linear operations such as smoothing, blurring, motion, \etc. Moreover, although we can also learn the size of each filter \wrt the linear operation $\Phi$, the irregular module will reduce the efficiency of computing units (\eg CPU and GPU). Thus, we suggest to let $d$ in a Ghost module be a fixed value and utilize depthwise convolution to implement Eq.~\ref{eq:ghost} for building highly efficient deep neural networks in the following experiments.

\begin{table}[htb]
	\centering
	\small
	\renewcommand\arraystretch{1.0}
	\caption{The performance of the proposed Ghost module with different $d$ on CIFAR-10.}
	\vspace{0.1em}
	\label{tab:p}
	\begin{tabular}{c||c|c|c}
		\hline
		$d$         &  Weights (M) & FLOPs (M) & Acc. (\%) \\ \hline\hline
		VGG-16         & 15.0    & 313    & 93.6     \\ 
		1         & 7.6   & 157  & 93.5     \\ 
		3       &    7.7  & 158 &    93.7 \\
		5       &   7.7  & 160   &  93.4  \\ 
		7        &   7.7  & 163    & 93.1  \\ \hline
	\end{tabular}
	\vspace{-1em}
\end{table}

\begin{table}[htb]
	\centering
	\small
	\renewcommand\arraystretch{1.0}
	\caption{The performance of the proposed Ghost module with different $s$ on CIFAR-10.}
	\vspace{0.1em}
	\label{tab:s}
	\begin{tabular}{c||c|c|c}
		\hline
		$s$      &  Weights (M)    &  FLOPs (M) & Acc. (\%) \\ \hline\hline
		VGG-16      & 15.0   & 313        & 93.6     \\ 
		2     & 7.7     & 158        & 93.7     \\ 
		3     & 5.2 & 107 &  93.4 \\
		4      & 4.0  &   80      &  93.0   \\
		5     & 3.3   &   65      &  92.9  \\
		\hline
	\end{tabular}
	\vspace{-1em}
\end{table}

\subsubsection{CIFAR-10.} We evaluate the proposed Ghost module on two popular network architectures, \ie VGG-16~\cite{vgg} and ResNet-56~\cite{resnet}, on CIFAR-10 dataset. Since VGG-16 is originally designed for ImageNet, we use its variant~\cite{cifar-vgg} which is widely used in literatures for conducting the following experiments. All the convolutional layers in these two models are replaced by the proposed Ghost module, and the new models are denoted as Ghost-VGG-16 and Ghost-ResNet-56, respectively. Our training strategy closely follows the settings in~\cite{resnet}, including momentum, learning rate, \etc. We first analyze the effects of the two hyper-parameters $s$ and $d$ in Ghost module, and then compare the Ghost-models with the state-of-the-art methods.

%

\paragraph{Analysis on Hyper-parameters.} As described in Eq.~\ref{eq:ghost}, the proposed Ghost Module for efficient deep neural networks has two hyper-parameters, \ie $s$ for generating $m = n/s$ intrinsic feature maps, and kernel size $d\times d$ of linear operations (\ie the size of depthwise convolution filters) for calculating ghost feature maps. The impact of these two parameters are tested on the VGG-16 architecture.

First, we fix $s=2$ and tune $d$ in $\{1,3,5,7\}$, and list the results on CIFAR-10 validation set in Table~\ref{tab:p}. We can see that the proposed Ghost module with $d=3$ performs better than smaller or larger ones. This is because that kernels of size $1\times 1$ cannot introduce spatial information on feature maps, while larger kernels such as $d=5$ or $d=7$ lead to overfitting and more computations. Therefore, we adopt $d=3$ in the following experiments for effectiveness and efficiency.

After investigating the kernel sizes used in the proposed Ghost module, we keep $d=3$ and tune the other hyper-parameter $s$ in the range of $\{2,3,4,5\}$. In fact, $s$ is directly related to the computational costs of the resulting network, that is, larger $s$ leads to larger compression and  speed-up ratio as analyzed in Eq.~\ref{eq:rc} and Eq.~\ref{eq:rs}. From the results in Table~\ref{tab:s}, when we increase $s$, the FLOPs are reduced significantly and the accuracy decreases gradually, which is as expected. Especially when $s=2$ which means compress VGG-16 by $2\times$, our method performs even slightly better than the original model, indicating the superiority of the proposed Ghost module.

\begin{table}[htb]
	\centering
	\small
	\renewcommand\arraystretch{1.0}
	\caption{Comparison of state-of-the-art methods for compressing VGG-16 and ResNet-56 on CIFAR-10. - represents no reported results available.}
	\vspace{0.1em}
	\label{tab:cifar10}
	\begin{tabular}{c||c|c|c}
		\hline
		Model         & Weights & FLOPs & Acc. (\%) \\ \hline\hline
		VGG-16   & 15M        & 313M        & 93.6     \\ 
		$\ell_1$-VGG-16~\cite{l1-pruning,rethinking-pruning} & 5.4M & 206M & 93.4 \\
		SBP-VGG-16~\cite{he2017channel} & - & 136M & 92.5 \\
		Ghost-VGG-16 ($s$=2) & 7.7M         &   158M      & \textbf{93.7}     \\ 
		\hline
		ResNet-56   & 0.85M         & 125M        & 93.0     \\
		CP-ResNet-56~\cite{he2017channel} & - & 63M & 92.0 \\ 
		$\ell_1$-ResNet-56~\cite{l1-pruning,rethinking-pruning} & 0.73M & 91M & 92.5 \\
		AMC-ResNet-56~\cite{amc} & - & 63M & 91.9 \\
		Ghost-ResNet-56 ($s$=2) & 0.43M         & 63M        & \textbf{92.7}     \\ \hline
	\end{tabular}
	\vspace{-1em}
\end{table}

\begin{table*}[htb]
	\centering
	\small
	\renewcommand\arraystretch{1.0}
	\caption{Comparison of state-of-the-art methods for compressing ResNet-50 on ImageNet dataset.}
	\vspace{0.1em}
	\label{tab:resnet50}
	\begin{tabular}{c||c|c|c|c}
		\hline
		Model         & Weights (M) & FLOPs (B) & Top-1 Acc. (\%) & Top-5 Acc. (\%) \\ \hline\hline
		ResNet-50~\cite{resnet}   & 25.6     & 4.1        & 75.3 &  92.2    \\\hline
		Thinet-ResNet-50~\cite{thinet} & 16.9 & 2.6 & 72.1 & 90.3  \\
		NISP-ResNet-50-B~\cite{nisp} & 14.4 & 2.3 & - & 90.8 \\ 
		Versatile-ResNet-50~\cite{versatile} & 11.0 &  3.0 & 74.5 & 91.8 \\
		SSS-ResNet-50~\cite{huang2018data} & - & 2.8 & 74.2 & 91.9 \\
		Ghost-ResNet-50 ($s$=2) & 13.0      & 2.2     &  \textbf{75.0} & \textbf{92.3}     \\
		\hline
		Shift-ResNet-50~\cite{shift} & 6.0 & - & 70.6 & 90.1  \\
		Taylor-FO-BN-ResNet-50~\cite{molchanov2019importance} & 7.9 & 1.3 & 71.7 & - \\
		Slimmable-ResNet-50 0.5$\times$~\cite{slimmable} & 6.9 & 1.1 & 72.1 & -  \\
		MetaPruning-ResNet-50~\cite{metapruning} & - & 1.0 & 73.4 & - \\
		Ghost-ResNet-50 ($s$=4) & 6.5   & 1.2  & \textbf{74.1} & \textbf{91.9}  \\ \hline
	\end{tabular}
	\vspace{-1em}
\end{table*}

\paragraph{Comparison with State-of-the-arts.} We compare Ghost-Net with several representative state-of-the-art models on both VGG-16 and ResNet-56 architectures. The compared methods include different types of model compression approaches, $\ell_1$ pruning~\cite{l1-pruning,rethinking-pruning}, SBP~\cite{he2017channel}, channel pruning (CP)~\cite{he2017channel} and AMC~\cite{amc}. For VGG-16, our model can obtain an accuracy slightly higher than the original one with a 2$\times$ acceleration, which indicates that there is considerable redundancy in the VGG model. Our Ghost-VGG-16 ($s=2$) outperforms the competitors with the highest performance ($93.7\%$) but with significantly fewer FLOPs. For ResNet-56 which is already much smaller than VGG-16, our model can achieve comparable accuracy with baseline with 2$\times$ speed-up. We can also see that other state-of-the-art models with similar or larger computational cost obtain lower accuracy than ours.

\begin{figure}[htb]
	\vspace{-0.5em}
	\centering
	\includegraphics[width=1.0\linewidth]{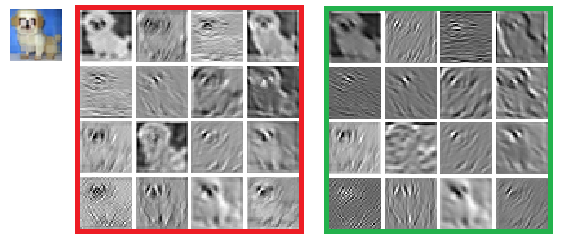}
	\caption{The feature maps in the 2nd layer of Ghost-VGG-16. The left-top image is the input, the feature maps in the left red box are from the primary convolution, and the feature maps in the right green box are after the depthwise transformation.}
	\label{fig:vis}
	\vspace{-1em}
\end{figure}

\paragraph{Visualization of Feature Maps.} 
We also visualize the feature maps of our ghost module as shown in Figure~\ref{fig:vis}. Although the generated feature maps are from the primary feature maps, they exactly have significant difference which means the generated features are flexible enough to satisfy the need for the specific task.

\begin{figure}[htb]
	\centering
	\includegraphics[width=0.85\linewidth]{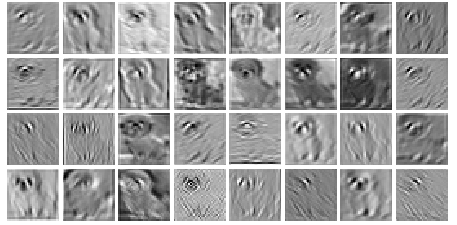}
	\caption{The feature maps in the 2nd layer of vanilla VGG-16.}
	\label{fig:vgg-vis}
	\vspace{-1em}
\end{figure}


\subsubsection{Large Models on ImageNet} 
We next embed the Ghost module in the standard ResNet-50~\cite{resnet} and conduct experiments on the large-scale ImageNet dataset. ResNet-50 has about 25.6M parameters and 4.1B FLOPs with a top-5 error of $7.8\%$. We use our Ghost module to replace all the convolutional layers in ResNet-50 to obtain compact models and compare the results with several state-of-the-art methods, as detailed in Table~\ref{tab:resnet50}. The training settings such as the optimizer, the learning rate, and the batch size, are totally the same as those in~\cite{resnet} for fair comparisons.

From the results in Table~\ref{tab:resnet50}, we can see that our Ghost-ResNet-50 ($s$=2) obtains about 2$\times$ acceleration and compression ratio, while maintaining the accuracy as that of the original ResNet-50. Compared with the recent state-of-the-art methods including Thinet~\cite{thinet}, NISP~\cite{nisp}, Versatile filters~\cite{versatile} and Sparse structure selection (SSS)~\cite{huang2018data}, our method can obtain significantly better performance under the 2$\times$ acceleration setting. When we further increase $s$ to 4, Ghost-based model has only a $0.3\%$ accuracy drop with an about 4$\times$ computation speed-up ratio. In contrast, compared methods~\cite{shift,slimmable} with similar weights or FLOPs have much lower performance than ours.

\subsection{GhostNet on Visual Benchmarks}
After demonstrating the superiority of the proposed Ghost module for efficiently generating feature maps, we then evaluate the well designed GhostNet architecture as shown in Table~\ref{tab:GhostNet} using Ghost bottlenecks on image classification and object detection tasks, respectively.

\subsubsection{ImageNet Classification} To verify the superiority of the proposed GhostNet, we conduct experiments on ImageNet classification task. We follow most of the training settings used in~\cite{shufflenet}, except that the initial learning rate is set to 0.4 when batch size is 1,024 on 8 GPUs. All the results are reported with single crop top-1 performance on ImageNet validation set. For GhostNet, we set kernel size $k=1$ in the primary convolution and $s=2$ and $d=3$ in all the Ghost modules for simplicity.

Several modern small network architectures are selected as competitors, including MobileNet series~\cite{mobilenet,mobilev2,mobilenetv3}, ShuffleNet series~\cite{shufflenet,shufflev2}, ProxylessNAS~\cite{proxylessnas}, FBNet~\cite{fbnet}, MnasNet~\cite{mnasnet}, \etc. The results are summarized in Table~\ref{tab:GhostNet}. The models are grouped into three levels of computational complexity typically for mobile applications, \ie $\sim$50, $\sim$150, and 200-300 MFLOPs. From the results, we can see that generally larger FLOPs lead to higher accuracy in these small networks which shows the effectiveness of them. Our GhostNet outperforms other competitors consistently at various computational complexity levels, since GhostNet is more efficient in utilizing computation resources for generating feature maps.

\begin{figure}[htb]
	\vspace{-0.5em}
	\centering
	\small
	\includegraphics[width=0.95\linewidth]{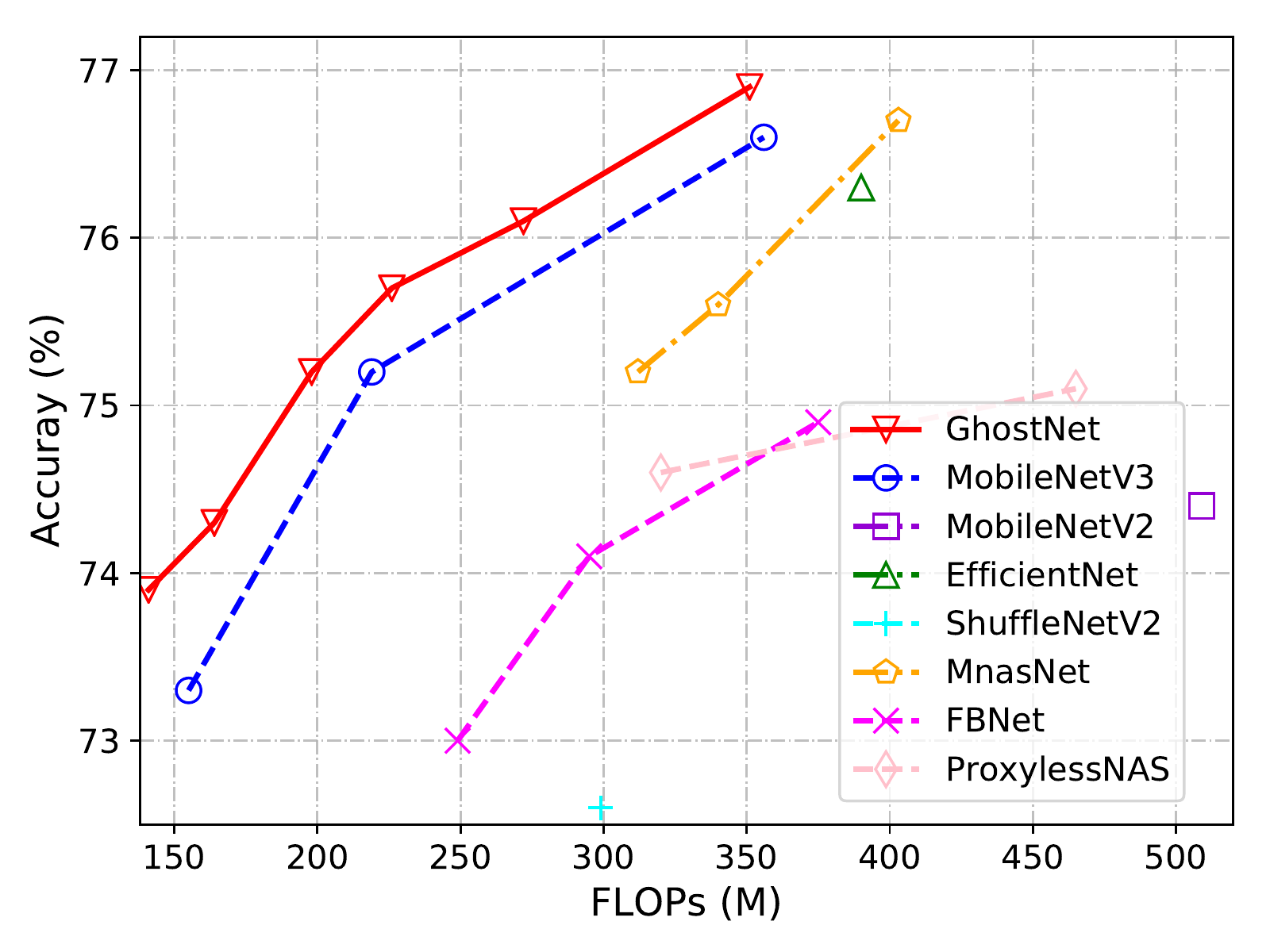}
	\vspace{-0.5em}
	\caption{Top-1 accuracy \emph{v.s.} FLOPs on ImageNet dataset.}
	\label{fig:speed}
	\vspace{-1em}
\end{figure}

\begin{table*}[htb]
	\centering
	\small
	\renewcommand\arraystretch{1.0}
	\caption{Comparison of state-of-the-art small networks over classification accuracy, the number of weights and FLOPs on ImageNet dataset.}
	\vspace{0.1em}
	\label{tab:GhostNet}
	\begin{tabular}{c||c|c|c|c}
		\hline
		Model         & Weights (M) & FLOPs (M) & Top-1 Acc. (\%) & Top-5 Acc. (\%) \\ \hline\hline
		ShuffleNetV1 0.5$\times$ (g=8)~\cite{shufflenet}   & 1.0   & 40     & 58.8 &  81.0  \\
		MobileNetV2 0.35$\times$~\cite{mobilev2}   &  1.7   &  59     &  60.3  & 82.9  \\
		ShuffleNetV2 0.5$\times$~\cite{shufflev2}   & 1.4   & 41     & 61.1  & 82.6  \\
		MobileNetV3 Small 0.75$\times$~\cite{mobilenetv3}   & 2.4  &  44     &  65.4   & -  \\
		GhostNet 0.5$\times$ &  2.6   &  42   &  \textbf{66.2}   & \textbf{86.6}   \\ 
		\hline
		MobileNetV1 0.5$\times$~\cite{mobilenet}   &  1.3   &  150     &  63.3  & 84.9   \\
		MobileNetV2 0.6$\times$~\cite{mobilev2,shufflev2}   & 2.2   &  141    & 66.7  & -\\
		ShuffleNetV1 1.0$\times$ (g=3)~\cite{shufflenet}   & 1.9   & 138     & 67.8  & 87.7   \\
		ShuffleNetV2 1.0$\times$~\cite{shufflev2}   & 2.3   & 146     & 69.4  & 88.9  \\
		MobileNetV3 Large 0.75$\times$~\cite{mobilenetv3}   & 4.0  &  155     &  73.3   & -  \\
		GhostNet 1.0$\times$ & 5.2    & 141  &  \textbf{73.9}  & \textbf{91.4}   \\ 
		\hline
		MobileNetV2 1.0$\times$~\cite{mobilev2}   & 3.5   & 300     & 71.8  & 91.0  \\
		ShuffleNetV2 1.5$\times$~\cite{shufflev2}   & 3.5   & 299     & 72.6  & 90.6   \\
		FE-Net 1.0$\times$~\cite{sparse-shift}   & 3.7   & 301     & 72.9  & -   \\
		FBNet-B~\cite{fbnet}    & 4.5   &  295   & 74.1  & -   \\
		ProxylessNAS~\cite{proxylessnas} & 4.1  &  320     &  74.6   &  92.2  \\
		MnasNet-A1~\cite{mnasnet}   & 3.9  &  312     &  75.2   &  92.5  \\
		MobileNetV3 Large 1.0$\times$~\cite{mobilenetv3}   & 5.4  &  219     &  75.2   & -  \\
		GhostNet 1.3$\times$ &  7.3    & 226   &  \textbf{75.7}  & \textbf{92.7}   \\ 
		\hline
	\end{tabular}
	\vspace{-1em}
\end{table*}

\begin{figure}[htb]
	\centering
	\small
	\includegraphics[width=0.95\linewidth]{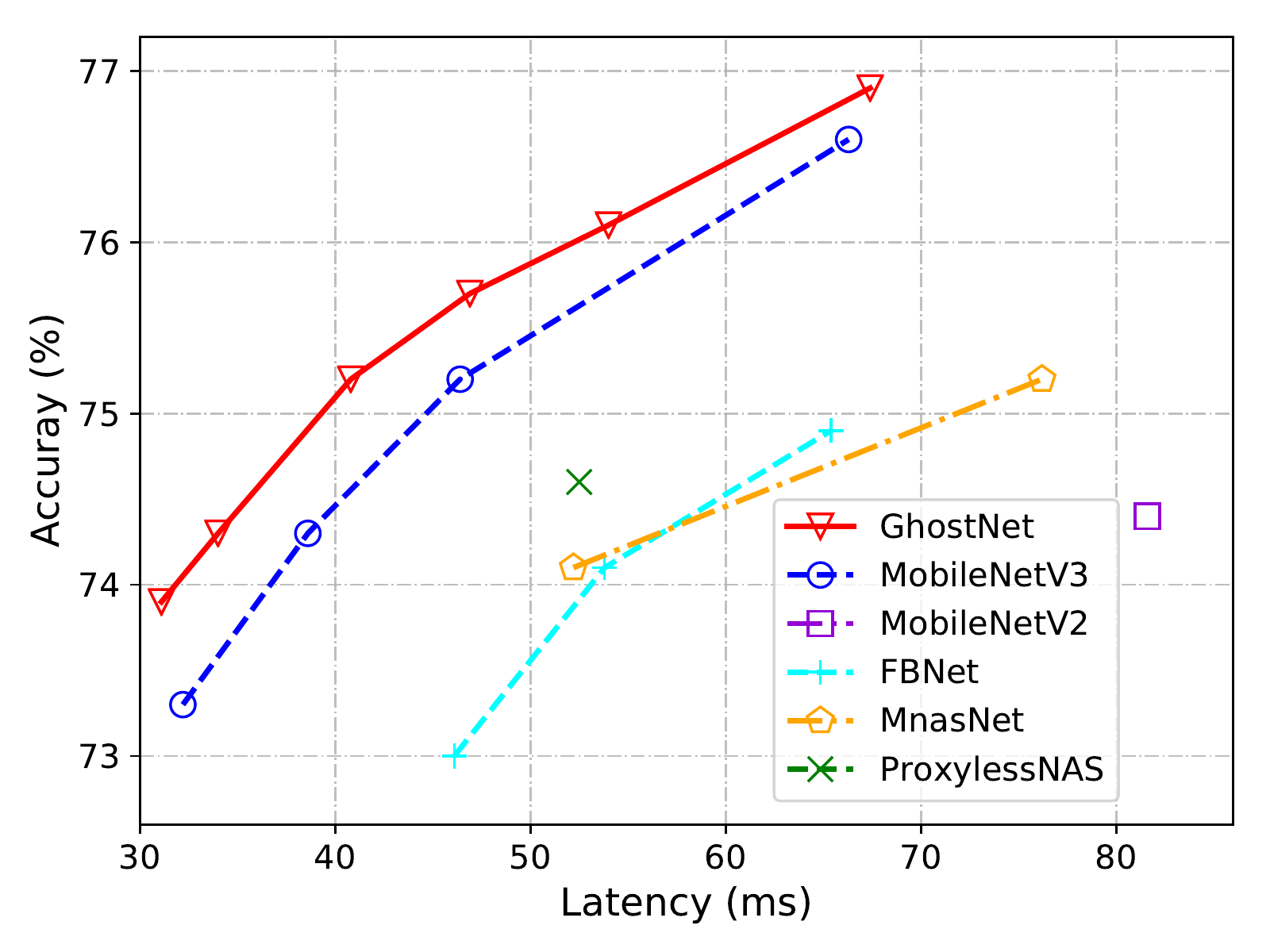}
	\vspace{-0.5em}
	\caption{Top-1 accuracy \emph{v.s.} latency on ImageNet dataset.}
	\label{fig:speed2}
	\vspace{-0.5em}
\end{figure}

\paragraph{Actual Inference Speed.} Since the proposed GhostNet is designed for mobile applications, we further measure the actual inference speed of GhostNet on an ARM-based mobile phone using the TFLite tool~\cite{tensorflow}. Following the common settings in ~\cite{mobilenet,mobilev2}, we use single-threaded mode with batch size 1. From the results in Figure~\ref{fig:speed2}, we can see that GhostNet obtain about 0.5\% higher top-1 accuracy than MobileNetV3 with the same latency, and GhostNet need less runtime to achieve similar performance. For example, GhostNet with 75.0\% accuracy only has 40 ms latency, while MobileNetV3 with similar accuracy requires about 45 ms to process one image. Overall, our models generally outperform the famous state-of-art models, \ie MobileNet series~\cite{mobilenet,mobilev2,mobilenetv3}, ProxylessNAS~\cite{proxylessnas}, FBNet~\cite{fbnet}, and MnasNet~\cite{mnasnet}.

\subsubsection{Object Detection} 
In order to further evaluate the generalization ability of GhostNet, we conduct object detection experiments on MS COCO dataset. We use the \emph{trainval35k} split as training data and report the results in mean Average Precision (mAP) on \emph{minival} split, following~\cite{fpn,retinanet}. Both the two-stage Faster R-CNN with Feature Pyramid Networks (FPN)~\cite{fasterrcnn,fpn} and the one-stage RetinaNet~\cite{retinanet} are used as our framework and GhostNet acts as a drop-in replacement for the backbone feature extractor. We train all the models using SGD for 12 epochs from ImageNet pretrained weights with the hyper-parameters suggested in~\cite{fpn,retinanet}. The input images are resized to a short side of 800 and a long side not exceeding 1333. Table~\ref{tab:voc2007} shows the detection results, where the FLOPs are calculated using $224\times224$ images as common practice. With significantly lower computational costs, GhostNet achieves similar mAP with MobileNetV2 and MobileNetV3, both on one-stage RetinaNet and two-stage Faster R-CNN frameworks.

\begin{table}[htb]
	\centering
	\small
	\renewcommand\arraystretch{1.0}
	\caption{Results on MS COCO dataset.}
	\vspace{0.1em}
	\label{tab:voc2007}
	\setlength{\tabcolsep}{4pt}{
	\begin{tabular}{c||c|c|c}
		\hline
		Backbone      & \tabincell{c}{Detection\\Framework}   & \tabincell{c}{Backbone\\FLOPs} & mAP \\ \hline\hline
		MobileNetV2 1.0$\times$~\cite{mobilev2}  & \multirow{3}{*}{RetinaNet}   &  300M    & 26.7\%  \\
		MobileNetV3 1.0$\times$~\cite{mobilenetv3}    &  &  219M    & 26.4\%  \\
		GhostNet 1.1$\times$ & & 164M & 26.6\%  \\\hline
		MobileNetV2 1.0$\times$~\cite{mobilev2}  & \multirow{3}{*}{Faster R-CNN}   &  300M    & 27.5\%  \\
		MobileNetV3 1.0$\times$~\cite{mobilenetv3}    &  &  219M    & 26.9\%  \\
		GhostNet 1.1$\times$ & & 164M & 26.9\%  \\\hline
	\end{tabular}
	}
	\vspace{-1em}
\end{table}

\section{Conclusion}\label{Conclusion}
To reduce the computational costs of recent deep neural networks, this paper presents a novel Ghost module for building efficient neural architectures. The basic Ghost module splits the original convolutional layer into two parts and utilizes fewer filters to generate several intrinsic feature maps. Then, a certain number of cheap transformation operations will be further applied for generating ghost feature maps efficiently. The experiments conducted on benchmark models and datasets illustrate that the proposed method is a plug-and-play module for converting original models to compact ones while remaining the comparable performance. In addition, the GhostNet built using the proposed new module outperforms state-of-the-art portable neural architectures, in both terms of efficiency and accuracy.

\section*{Acknowledgment}
We thank anonymous reviewers for their helpful comments. Chang Xu was supported by the Australian Research Council under Project DE180101438.

{\small
\bibliographystyle{ieee_fullname}
\bibliography{ref}
}

\end{document}